\documentclass[runningheads]{llncs}

\usepackage{eccv}

\usepackage{appendix}
\usepackage{eccvabbrv}

\usepackage{graphicx}
\usepackage{booktabs}
\usepackage{multirow}

\usepackage[accsupp]{axessibility}  % Improves PDF readability for those with disabilities.

\usepackage{hyperref}

\usepackage{orcidlink}
\usepackage{graphicx}
\usepackage[ruled,vlined]{algorithm2e}
\usepackage{algorithmic}
\usepackage{xcolor}
\usepackage[table]{xcolor}
\usepackage{mathtools}
\usepackage{comment}
\usepackage{multirow}
\begin{document}

% ---------------------------------------------------------------
% TODO REVIEW: Replace with your title
\title{RADIANCE: Relative Adaptive Denoising with IP-Adapter for \\
Novel Concept Enhancement}
\newcommand{\abbrev}{RADIANCE\xspace}
% TODO REVIEW: If the paper title is too long for the running head, you can set
% an abbreviated paper title here. If not, comment out.
\titlerunning{RADIANCE}

% TODO FINAL: Replace with your author list. 
% Include the authors' OCRID for the camera-ready version, if at all possible.
%\author{First Author\inst{1}\orcidlink{0000-1111-2222-3333} \and
%Second Author\inst{2,3}\orcidlink{1111-2222-3333-4444} \and
%Third Author\inst{3}\orcidlink{2222--3333-4444-5555}}

\author{Zi-Xiang Ni\inst{1}\orcidlink{0009-0002-4123-5132}\and 
Bo-Lun Huang\inst{1}\orcidlink{0009-0003-9804-5046}\and 
Teng-Fang Hsiao\inst{1}\orcidlink{0009-0004-4075-316X} \and 
Bo-Kai Ruan\inst{1}\orcidlink{0000-0002-9847-3628} \and 
Hong-Han Shuai\inst{1}\orcidlink{0000-0003-2216-077X}\thanks{Corresponding author.}}
% TODO FINAL: Replace with an abbreviated list of authors.
%\authorrunning{F.~Author et al.}
% First names are abbreviated in the running head.
% If there are more than two authors, 'et al.' is used.
\authorrunning{Z.-X.~Ni et al.}
% TODO FINAL: Replace with your institution list.
%\institute{Princeton University, Princeton NJ 08544, USA \and
%Springer Heidelberg, Tiergartenstr.~17, 69121 Heidelberg, Germany
%\email{lncs@springer.com}\\
%\url{http://www.springer.com/gp/computer-science/lncs} \and
%ABC Institute, Rupert-Karls-University Heidelberg, Heidelberg, Germany\\
%\email{\{abc,lncs\}@uni-heidelberg.de}}
\institute{National Yang Ming Chiao Tung University, Hsinchu, Taiwan\\
\email{\{zaq312.ee12, kevin503.ee12, bluedyee.ee09, bkruan.ee11, hhshuai\}@nycu.edu.tw}}

\maketitle

\begin{abstract}
Text-to-image (T2I) diffusion models have achieved striking progress but still struggle to synthesize rare concepts involving unusual attribute-object pairings, often resulting in concept omission or semantic drift where a dominant entity overwhelms the generation. Tracing these failures to a lack of compositional balance during the denoising trajectory, we propose RADIANCE, a training-free framework that treats inference as a closed-loop feedback process. RADIANCE augments pretrained backbones with three modular components: (1) a Compositional Similarity Monitor (CSM) that tracks the emergence of objects and attributes in intermediate latents via CLIP-based feedback; (2) a Bidirectional Scale Controller (BSC) that applies a reactive "restoring force" using positive and negative IP-Adapter scales to rebalance biased trajectories; and (3) a Feedback Guidance Scheduler (FGS) that coordinates these updates across timesteps without additional training. We further extend the framework to multi-object prompts via Delayed Adapter Activation (DAA) and Layer-wise Alternating Guidance (LAG) to prevent premature concept fusion. By overlapping monitoring and denoising through pipelined execution, RADIANCE maintains competitive latency while significantly enhancing the per-sample success rate and effective throughput. Experiments on RareBench and T2I-CompBench demonstrate that RADIANCE consistently enhances compositional alignment and perceptual quality over state-of-the-art baselines.
%\keywords{Diffusion Models · Image Generation · Training Free}
\end{abstract}    
\section{Introduction}

Text-to-image (T2I) diffusion models have rapidly become a foundation for visual content creation, enabling users to synthesize images directly from natural language prompts~\cite{ho2020denoising,pixart,dhariwal2021diffusion,sd1.5,ding2021cogview, chen2025janus, showo, emu2}. By combining with powerful text encoders such as CLIP~\cite{clip}, modern systems~\cite{nichol2021glide,saharia2022photorealistic,sdxl} can produce diverse and photorealistic samples that follow user intent. These capabilities have unlocked applications in digital art and design, advertising, and data augmentation for downstream vision tasks~\cite{ACMComputingSurvey24}. Recent MM-DiT-based architectures \eg, SD~3.5~\cite{sd3}, further improve scalability and cross-modal reasoning, narrowing the gap between model outputs and human imagination.

\begin{figure}[t]

    \centering
    \captionsetup{type=figure}
    \includegraphics[width=\linewidth]{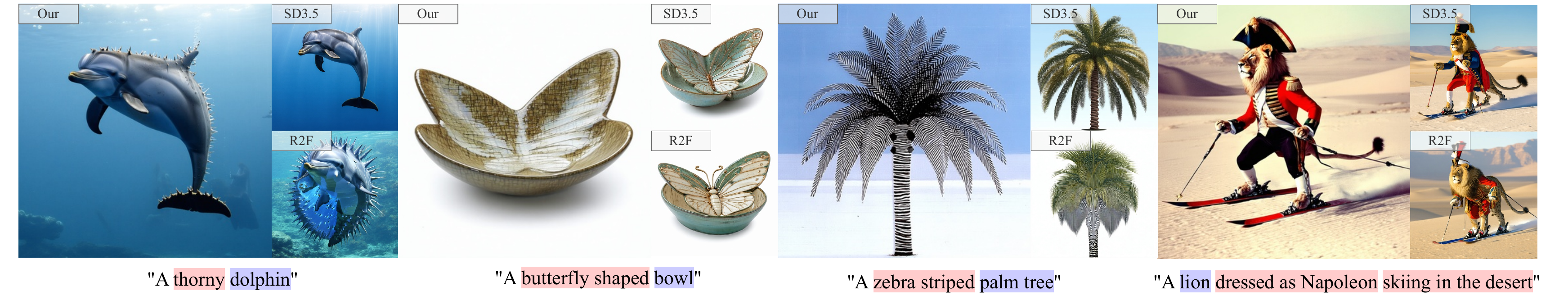}
    \captionof{figure}{\textbf{Generated samples from \abbrev across different rare concept prompts (attributes highlighted in \textcolor{red!50}{red} and objects in \textcolor{blue!50}{blue}}. Prior methods typically focus on either generating objects or capturing attributes, but struggle to merge both faithfully. For example, as shown in the first column, existing approaches~\cite{sd3, r2f} either fail to generate thorns or lose the dolphin semantics. In contrast, our training-free approach, \abbrev, yields coherent compositions that correctly combine rare attributes with their corresponding objects.}
    \label{fig:teaser}
    \vspace{-12pt}
\end{figure}

However, even the strongest T2I diffusion models still struggle when prompts describe \emph{rare concepts}, such as objects with unusual attributes, shapes, or textures (\eg, ``a thorny dolphin'' or ``a zebra striped palm tree'' in Fig.~\ref{fig:teaser}). In practice, training data follow a long-tailed distribution: frequent attribute-object combinations are well represented, while rare concepts appear only a handful of times~\cite{huang2023composer,r2f}. As a result, models tend to fall back to frequent co-occurrences, which leads to attribute omission, entangled attributes, or incorrect bindings. Prior work on compositional and attribute aware generation~\cite{chefer2023attend,li2023gligen,huang2023composer,lian2023llm,zarei2024improving,butt2024colorpeel,laria2025leveraging,huang2025t2i, hsiao2025tf} improves how colors, shapes, and parts attach to a base object, but mainly for common concepts. More recent training-free strategies such as Rare to Frequent (R2F)~\cite{r2f} and prompt editing methods~\cite{justin_rap} replace rare tokens with semantically related frequent proxies and then switch back to the original prompt at selected timesteps, partially recovering the presence of rare concepts.

Despite the advances, existing approaches exhibit characteristic failure modes for rare concept generation. The rare attribute often materializes as disjoint entity in the image (\eg, for the prompt \textit{``a butterfly shaped bowl''} the model produces a normal bowl and a separate butterfly instead of a bowl whose silhouette forms a butterfly in Fig.~\ref{fig:teaser}, or the base object loses its defining characteristics and becomes distorted (\eg, for \textit{``a thorny dolphin''} the dolphin collapses into a spiky creature that no longer looks like a dolphin in Fig.~\ref{fig:teaser}. In both cases, the semantic balance between the rare attribute and the base object is broken. Prompt-level methods such as R2F~\cite{r2f} improves rare word expression but does not monitor object-attribute alignment, failing to correct step-wise drift when one component dominates. Other controllable generation methods~\cite{initno,rassin2023linguistic} use \textit{static} guidance scales, making them less adaptive to prompt dynamics.

To resolve these compositional failures, we introduce \textbf{RADIANCE} (\underline{R}are \underline{A}ttribute \underline{Di}ffusion with \underline{A}daptive \underline{N}ovel \underline{C}oncept \underline{E}nhancement), a training-free framework that reframes inference as a closed-loop feedback process. RADIANCE augments pretrained T2I backbones with three lightweight modules operating entirely at inference time. First, the Compositional Similarity Monitor (CSM) utilizes CLIP to extract real-time similarity signals from intermediate latents to track the emergence of both the base object and its rare attributes. Second, the Bidirectional Scale Controller (BSC) translates these signals into a reactive "restoring force"; it applies positive or negative IP-Adapter scales to either reinforce lagging concepts or actively suppress dominant ones, thereby maintaining semantic equilibrium. Finally, the Feedback Guidance Scheduler (FGS) coordinates these scale updates across the denoising trajectory, ensuring the rare attribute is faithfully integrated while preserving the base object's identity. Together, these components form a robust controller that rebalances concept influence online without modifying the underlying diffusion model.

Furthermore, building on this single-object controller, \abbrev~extends to prompts with multiple rare objects: it first synthesizes attribute-faithful single-object images as references and then uses them as guidance for multi-object generation with \textit{delayed adapter activation} and \textit{layer-wise alternating guidance}, which prevents premature fusion and preserves object identities in the final composition. The contributions of this work can be summarized as follows:
\begin{itemize}
    \item We propose \abbrev, a training-free method that combines similarity monitoring, bidirectional scaling, and an adaptive guidance scheduler to maintain balance between rare attributes and base objects along the denoising trajectory for single object prompt.

    \item We further extend \abbrev to multi-object rare compositions by first synthesizing attribute faithful single object reference images and then using delayed and layer-wise guidance, which prevents premature fusion and preserves distinct object identities in the final composition.

    \item Experiments on RareBench and T2I-CompBench manifest that~\abbrev significantly improves compositional alignment and visual fidelity over SOTA baselines. These gains are rigorously validated through automated scoring and extensive human preference studies.

\end{itemize}

\section{Preliminary}
\subsection{Problem Formulation}
\label{sec:problem}
Let $c$ denote a text prompt. We decompose $c$ into one object term $o$ and $m$ attribute terms \(A=\{a_i\}_{i=1}^{m}\), where each term is a word or a short phrase and $A$ is treated as an unordered set and $m \in \mathbb{N}_{\ge 0}$. The goal of compositional text-to-image generation is to produce an image $x_0$ that depicts the object $o$ while satisfying the attributes $\{a_i\}_{i=1}^{m}$ as faithfully as possible. For example, $o$ may be \textit{``monkfish''} and the attributes may include \textit{``horned''} and \textit{``bearded''}.

\paragraph{Rare Concepts.}
A concept \(c\) is \emph{rare} if its probability under the training distribution is near zero:
\[
p_{\mathrm{data}}(c)\approx 0,\qquad c\in\mathcal{C}_R\subseteq\mathcal{C}.
\]
While a model may possess sufficient priors for individual components (\textbf{atomic concepts}), the \textbf{joint distribution} of their pairing is often poorly represented, defined here as \textbf{compositional rarity} where $P_{data}(c) \approx 0$. This data sparsity induces a ``prior dominance'' effect during sampling: the high-frequency prior of the base object (e.g., a standard dolphin) effectively suppresses the lower-frequency rare attribute (e.g., thorns). Consequently, the denoising trajectory gravitates toward frequent co-occurrences, leading to attribute omission or semantic entanglement even when the model is capable of synthesizing the individual components in isolation~\cite{t2i_compbench, r2f}.

%We will later monitor object–attribute faithfulness with similarity scores and maintain balance throughout the denoising trajectory (Sec.~\ref{sec:method}), but the optimization happens at inference time and requires no additional training.

\subsection{Related Work}

\subsubsection{Text-to-Image Models}
Diffusion models have become the dominant framework for T2I generation, achieving remarkable visual fidelity and strong semantic alignment~\cite{ho2020denoising,pixart,dhariwal2021diffusion,sd1.5,ding2021cogview,chen2025janus,showo,emu2}. For example, Latent Diffusion Models (LDMs)~\cite{sd1.5,sdxl,pixart} pioneered the integration of CLIP-based semantic conditioning with the efficiency of a VAE latent space~\cite{kingma2014vae}. 
More recent architectures based on MM-DiT, such as Stable Diffusion 3.5~\cite{sd3} and FLUX~\cite{flux2024}, extend scalability and cross-modal reasoning by incorporating larger language models like T5~\cite{t5}, thereby narrowing the gap between model outputs and human imagination. Despite the advancements, generating coherent and faithful images for rare or highly compositional prompts remains challenging, as models tend to overfit to frequent co-occurrences and deviate from the intended composition~\cite{chefer2023attend,r2f,mou2024t2i}.

\subsubsection{Rare Concept and Compositional Generation}
Growing research focuses on attribute binding and compositional control in diffusion models~\cite{chefer2023attend,li2023gligen,huang2023composer,lian2023llm,zarei2024improving,butt2024colorpeel,laria2025leveraging,huang2025t2i}, aiming to better associate attribute with a base object.
However, these methods perform poorly on rare concepts due to the long-tailed nature of web-scale data, which biases models toward frequent patterns and underrepresents rare ones~\cite{r2f}.
Training-free approaches such as R2F~\cite{r2f} address this issue by substituting rare tokens with semantically related proxies and reverting to the original prompt at specific timesteps.
\textbf{Yet, they do not explicitly correct the balance between the rare attribute and its base object during denoising.}
In contrast, our method treats inference as a feedback process—continuously measuring object–attribute alignment and adaptively rebalancing their influence for more coherent and accurate generation.

%\paragraph{Image prompt adapters and IP-Adapter.} IP-Adapter~\cite{ye2023ip-adapter} enables pretrained T2I diffusion models~\cite{rombach2022high,saharia2022photorealistic} to ingest image prompts through a decoupled cross-attention pathway, preserving the original weights while adding an image-conditioned branch. Subsequent works explore related adapters and their use for controllable generation~\cite{zhang2023adding,mou2024t2i}, including UNIC-Adapter, which unifies image and instruction conditioning in a multi-modal transformer~\cite{duan2024unicadapter}, and IP-Composer, which composes multiple reference images into new visual concepts~\cite{dorfman2025ipcomposer}. These approaches mainly focus on expanding the range of conditioning signals and improving general control. Our work instead uses IP-Adapter as an actuator inside a feedback controller, \textit{i.e.}, we observe that both positive and negative scales on the image adapter branch are effective, and we exploit this bidirectional scaling to counter over-dominance of reference images and maintain compositional balance between rare attributes and base objects during sampling.

\section{\abbrev}
\subsection{Overview}
The composition of rare attribute-object pairs frequently fails as the sampling process gravitates toward familiar correlations, leading to either attribute omission or the loss of base object identity. To resolve this imbalance, we propose~\abbrev, which reframes inference as a feedback-driven process that continuously monitors and adjusts compositional alignment. Equipped with a flow-based backbone, the framework utilizes a Compositional Similarity Monitor (CSM) to decode intermediate latents and evaluate the emergence of each concept via CLIP-based similarity signals. Based on this real-time feedback, the Bidirectional Scale Controller (BSC) applies positive or negative scaling to the IP-Adapter pathway, providing a reactive ``restoring force'' that reinforces lagging components or suppresses dominant ones. This capability for negative scaling is especially critical, as it allows the model to actively counter excessive attribute influence rather than relying on additive guidance. Finally, the Feedback Guidance Scheduler (FGS) orchestrates the temporal policy, prioritizing the initial emergence of both concepts before shifting focus toward preserving object structural identity as the generation converges.

%Composing rare attribute–object pairs often fails because the sampler gravitates to familiar correlations, so either the attribute overwhelms the object or the object loses identity. To address this issue, we propose \abbrev. Equipped with the flow model~\cite{liu2023flow}, \abbrev performs inference as a feedback-driven process that continuously monitors and adjusts compositional balance. At each denoising step, the current latent is decoded and evaluated by the \textit{Compositional Similarity Monitor (CSM)} (\cref{sec:detection}), which measures how strongly the object and each attribute are expressed using CLIP-based similarity signals. The \textit{Bidirectional Scale Controller (BSC)} (\cref{sec:bias_correction}) then acts on this feedback, applying positive or negative scaling to the IP-Adapter pathway to either reinforce weak components or suppress dominant ones. Negative scaling is especially crucial, enabling the model to counter excessive attribute influence rather than merely adding guidance. Finally, the \textit{Feedback Guidance Scheduler (FGS)} (\cref{sec:adaptive_guidance}) governs the temporal policy, allowing both object and attribute emergence early in sampling before gradually shifting focus to preserving object identity as details converge.

\begin{figure*}[t]
    \centering
    \includegraphics[width=1\linewidth]{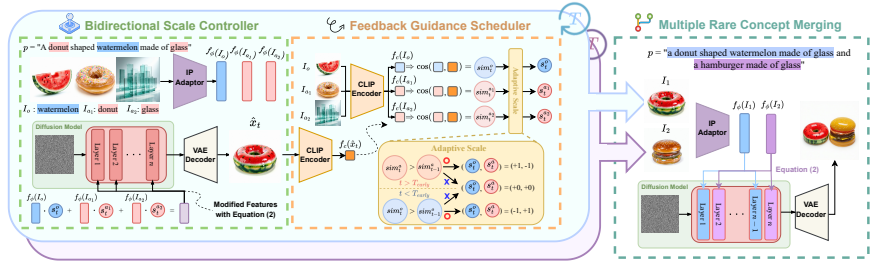}
    \caption{
    \textbf{Overview of the \abbrev pipeline.} 
    At each denoising step, the \textcolor{brown!60}{\textit{Feedback Guidance Scheduler}}~(\cref{sec:adaptive_guidance}) computes adaptive guidance. 
    The resulting scale is then applied by the \textcolor{green!60!black}{\textit{Bidirectional Scale Controller}}~(\cref{sec:bias_correction}) to inject reference information into the diffusion process. 
    Finally, multiple rare-concept outputs are merged via the \textcolor{cyan!60!black}{\textit{Multiple Rare Concepts Merging}}~(\cref{sec:multiobj}) that composed of DAA and LAG. 
    Together, these components ensure coherent and accurate generation of rare concepts.
    }
    \label{fig:method_flowchart}
    \vspace{-8pt}
\end{figure*}

This feedback loop operates without additional training or gradients and requires only a single decode per step, ensuring both efficiency and practicality.
For multi-object prompts (\cref{sec:multiobj}), we extend the same principle by activating multiple IP-Adapter branches layer-wise. Generated reference images of rare concepts are injected at different layers, allowing each object to first form independently before gradually interacting in later stages.
By unifying measurement, actuation, and scheduling within a single controller, \abbrev~preserves semantic balance and prompt fidelity throughout sampling, enabling coherent generation for both single- and multi-object rare compositions without retraining. Fig.~\ref{fig:method_flowchart} shows the model architecture of the proposed \abbrev.

%In practice, if the object weakens while an attribute grows, the next step boosts the object scale and applies a small negative scale to that attribute; if the attribute weakens while the object dominates, it flips these adjustments; if both remain stable, it keeps scales neutral. 

%definition and theoretical toy example
%detection with clip score
\subsection{Compositional Similarity Monitor (CSM)}
\label{sec:detection}
To track compositional alignment during inference, we introduce the \textit{Compositional Similarity Monitor (CSM)}, which measures how object and attribute contributions evolve at each denoising step. CSM requires a signal that is both lightweight, \ie, allowing stepwise computation, and reliable enough to indicate whether the current generation is dominated by the object or the attribute.
Instead of employing time-consuming VQA-style evaluation~\cite{manas2024improving}, we adopt CLIP~\cite{clip} as the backbone for similarity measurement, providing an efficient and scalable foundation for real-time compositional feedback. 

Specifically, for each prompt, we prepare one reference image $I_o$ for the base object and one reference image $I_a$ per attribute.\footnote{In practice, each object and attribute reference (for example, \textit{zebra stripped}, \textit{stone}) is obtained by a brief synthesized by the original backbone.} These reference images serve two roles. (1) they define the visual targets used for per-step similarity measurement. (2) they can serve as conditions for multiple image prompts in later stages.  During sampling, at each denoising step $t$, the diffusion backbone produces an intermediate latent $\hat{\mathbf{z}_t}$. We estimate the latent $\hat{\mathbf{z}}_t$ at time $t=0$ based on the flow formulation~\cite{liu2023flow}, then decode it into the current estimated image $\hat{\mathbf{x}}_t = \text{Decode}(\hat{\mathbf{z}}_t)$.
At each step $t$, we respectively compute the CLIP-based cosine similarity between $\hat{\mathbf{x}}_t$ and the corresponding reference images $I_{a_i}$ (for attributes) and $I_o$ (for the object) as follows:

\begin{equation}
\begin{aligned}
\text{sim}^{a_i}_t &= \cos(f_c(\hat{\mathbf{x}}_t), f_c(I_{a_i})), \\
\text{sim}^o_t &= \cos(f_c(\hat{\mathbf{x}}_t), f_c(I_{o})),
\end{aligned}
\label{eq:clip_sim}
\end{equation}
where $f_c$ denotes the CLIP image encoder.
This monitoring process requires only a single decoding operation per step and introduces no modification or fine-tuning to the backbone model.

\begin{figure}[t]
    \centering
    \includegraphics[width=1\linewidth]{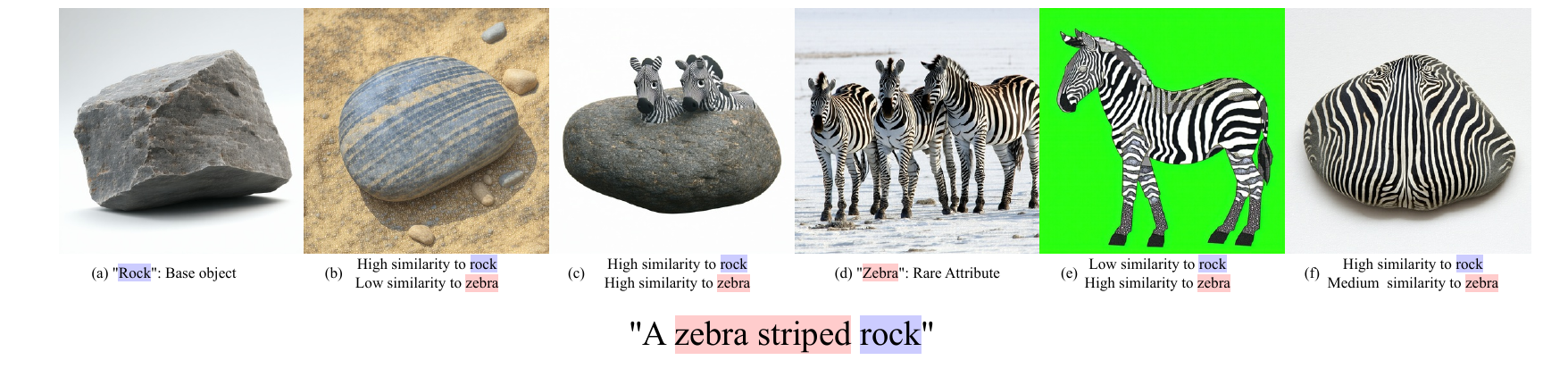}
    \vspace{-15pt}
    \caption{\textbf{Visualization of reference images and compositional outcomes for the prompt ``A \textcolor{red!50}{zebra striped} \textcolor{blue!50}{rock}''}. The left column shows reference images for the object (\textcolor{blue!50}{rock}) and attribute (\textcolor{red!50}{zebra striped}), while the right columns present three common failure cases and one correctly composed result.} 
    \label{fig:sim_bias}
    \vspace{-12pt}
\end{figure}

%%Here
As shown in \cref{fig:sim_bias}, leveraging CLIP-based similarity allows us to effectively distinguish different types of generation outcomes. 
In \cref{fig:sim_bias}(a) and (c), the generated images are dominated by a single aspect, representing common failure modes. 
In contrast, \cref{fig:sim_bias}(b) exhibits high $\mathrm{sim}^{o}_0$ and $\mathrm{sim}^{a_1}_0$ scores, yet the concepts merely co-exist rather than integrate, resulting in an implausible composition. 
Meanwhile, \cref{fig:sim_bias}(d) shows both similarities at moderate, balanced levels, indicating that the attribute is coherently expressed on the object. 
\textbf{These patterns underscore the core challenge: maintaining a proper balance between the rare attribute and the base object is more crucial than simply maximizing both similarity scores.}

By tracking $\mathrm{sim}^{a_i}_t$ and $\mathrm{sim}^{o}_t$ throughout the denoising trajectory, 
CSM can detect when the generation becomes dominated by one component (\eg, a rare attribute) 
or when another lags behind (\eg, the object), 
thereby providing the signal for adaptive guidance adjustment.

%negative ip-adapter scale
\subsection{Bidirectional Scale Controller (BSC)}
\label{sec:bias_correction}
The Compositional Similarity Monitor tells us when the denoising trajectory becomes biased. \textbf{Instead of forcing the similarities in Eq.~\eqref{eq:clip_sim} to match some preset numeric targets, which would vary with the prompt, timestep, and sampling noise, we only require that their trajectory remains balanced,} \ie, both the object and attribute similarities stay within a reasonable range and neither collapses nor dominates for steps. When the signals indicate this balance is broken at step $t$, we adjust the image-prompt pathway.

To this end, we propose \textit{Bidirectional Scale Controller (BSC)} based on IP-Adapter~\cite{ye2023ip-adapter}, which integrates image prompts into a pre-trained text-to-image diffusion model $\theta$ through a decoupled cross-attention mechanism. Given a text prompt $c$, a reference image $I$, an image encoder $f_{\phi}(\cdot)$, and a text encoder $f_{\mathcal{E}}(\cdot)$, IP-Adapter modifies the intermediate feature propagation of the model $\theta$ as:
\begin{equation}
\mathbf{h}_t^{l+1} = F_A^l\!\left(\mathbf{h}_t^l, f_\mathcal{E}(c)\right) 
                  + \sum_{i=1}^{n} s_i \, F_A^l\!\left(\mathbf{h}_t^l, f_\phi(I_{a_i})\right),
\end{equation}
where $\mathbf{h}_t^l$ denotes the intermediate feature of latent $\mathbf{z}_t$ at timestep $t$ and layer $l$, and $F_A^l(\cdot,\cdot)$ represents the attention block at layer $l$~\cite{attention_is_all_you_need_20kcites!}. The first argument provides the queries, while the second provides keys and values. In our setting, we attach one IP-Adapter branch for the base object and each attribute. For a concept with multiple reference images ${I_{a_1}, \dots, I_{a_n}}$, we weight the contribution from each reference image by a corresponding scalar $s_i$ and sum them, as shown in the second term, which can also be referred in the left part of \cref{fig:method_flowchart}.

\begin{figure}[t]
    \centering
    \includegraphics[width=1\linewidth]{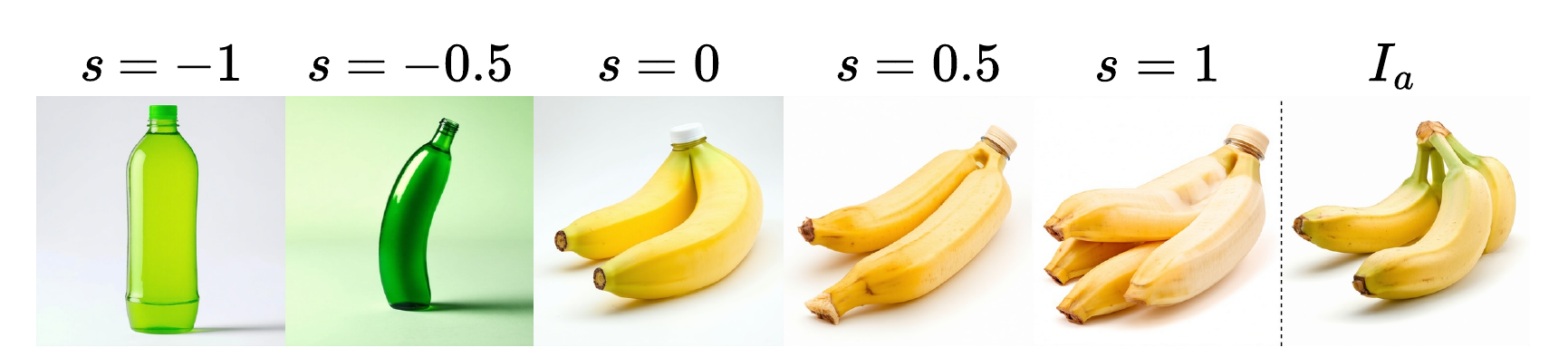}
    \caption{
    \textbf{Illustration of IP-Adapter guidance with different scales for the prompt ``A \textcolor{red!50}{banana shaped} \textcolor{blue!50}{bottle}''.}
    Here, the scalar $s$ controls the guidance of the IP-Adapter for the specific concept ``\textcolor{red!50}{banana shaped} '' ($I_a$).
    In this example, when generating without guidance from $I_a$ ($s=0$), the output deviates from the ``\textcolor{blue!50}{bottle}''.
    Interestingly, when setting $s$ to negative values ($-0.5$ and $-1$), the concept of ``\textcolor{red!50}{banana shaped}'' ($I_a$) is suppressed, leading to a more accurate depiction of the desired concept.
    }
    \label{fig:negative}
    \vspace{-12pt}
\end{figure}

One key empirical observation is that \textbf{IP-Adapter guidance remains effective even when its scale is negative}. Positive scales strengthen the influence of a reference image, while negative scales actively suppress it, as shown in \ref{fig:negative}. This bidirectional effect enables a simple bias–correction rule: when the generation becomes overly influenced by one component, we assign a negative scale to that component and a positive scale to the weaker one. The result is a reactive ``restoring force'' that pulls the trajectory back toward balanced composition.

\subsection{Feedback Guidance Scheduler (FGS)}
\label{sec:adaptive_guidance}

Building on the idea of negative-scale correction, we now integrate detection, decision, and correction into a single adaptive guidance pipeline, namely, Feedback Guidance Scheduler (FGS). As shown in Fig.~\ref{fig:method_flowchart}, FGS requires no additional training or gradient computation, 
and adjusts the IP-Adapter scales online according to compositional feedback provided by the CLIP-based detector.

\begin{figure}[t]
    \centering
    \includegraphics[width=0.6\linewidth]{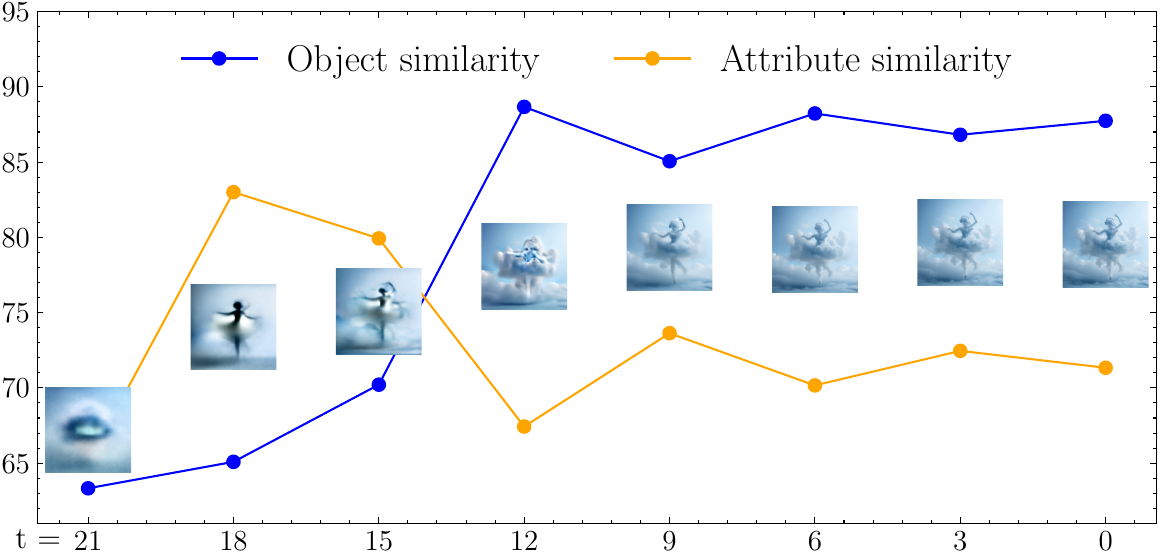}
    \caption{\textbf{Step-wise visualization of adaptive guidance.} 
    We shows each $\hat{x}_t$ and corresponding object, attribute similarity.}
    \label{fig:adaptive}
    \vspace{-10pt}
\end{figure}

\noindent\textbf{Adaptive Scale Assignment.} 
At each denoising step $t$, we obtain the current synthesized image $\hat{x}_t$ and compute the similarities
$\text{sim}^a_t$ and $\text{sim}^o_t$ with respect to the rare attribute and the base object, respectively. $\text{sim}^a_t$ and $\text{sim}^o_t$ are then compared with the similarities from the previous step, $\text{sim}^a_{t-1}$ and $\text{sim}^o_{t-1}$ to check whether either component is becoming over- or under-represented. If an imbalance is detected, we update the IP-Adapter scales at the next step so that the latent trajectory is nudged back toward a balanced composition. Fig.~\ref{fig:adaptive} provides a stepwise visualization of this process. 

For clarity, we first describe the policy for a single rare attribute $a$. Based on the similarity trends between two consecutive steps, we set the IP-Adapter guidance scales for the next step, $\big(s^a_{t-1}, s^o_{t-1}\big)$, as:
\begin{equation}
\label{eq:feedback_logic}
(s^a_{t-1}, s^o_{t-1}) =
\begin{cases}
(1, -1), & \text{if } t \ge T_{\text{early}} \ \text{and}\  \mathrm{sim}^a_t < \mathrm{sim}^a_{t+1} \\[1.5ex]
(-1, 1), & \text{else if } \mathrm{sim}^o_t < \mathrm{sim}^o_{t+1} \\[1.5ex]
(0, 0),  & \text{otherwise.}
\end{cases}
\end{equation}
Here, $T_{\text{early}}$ splits the trajectory into an early stage, with both the rare attribute and base object still forming, and a later stage, where the attribute is already stabilized and preserving object identity becomes the priority.
When $t \ge T_{\text{early}}$ and the attribute similarity keeps decreasing, indicating that the object becomes dominant, we set $(s_a, s_o) = (1, -1)$ to slightly suppress the object and reinforce the attribute.
Conversely, if the object similarity is the one that continues to decrease, we flip the signs to preserve the object instead.
When neither similarity exhibits a concerning trend, the scales remain $(0, 0)$, and the IP-Adapter functions as a neutral controller. Detailed comparison rules and the multi-attribute scheduler are provided in Appendix~\ref{sec:suppl_algorithm}. By continuously monitoring and correcting compositional drift in real-time, this ensures the faithful coexistence of rare and frequent concepts at zero additional training cost.

\subsection{Multiple Rare Concepts Merging}
\label{sec:multiobj}
%An overview of the full pipeline, from single object reference generation to multi object refinement, is shown in \cref{fig:method_flowchart}.
Building on our single rare concept framework, we further extend \abbrev~to prompts that contain multiple rare objects with their own attributes. Given a prompt $\mathcal{T} = \{o_1, o_2, \dots, o_n\}$ with $n$ rare objects, we first generate single-object images $I_{i}$ for each object $o_i$. We then reuse them as image prompts to steer a joint multi-object synthesis via IP-Adapter. \textbf{A major challenge arises if we simply feed all reference images into the IP-Adapter in parallel}: the model tends to mix up distinct objects. For example, a prompt such as ``a horned elephant and a hairy frog'' may produce a single object that mixes features of both the elephant and the frog, instead of two clearly separated objects. To mitigate premature fusion of multiple rare objects, we introduce two complementary strategies.

\noindent\textbf{Delayed Adapter Activation (DAA).} We disable IP Adapter guidance for the first $T_{\text{delay}}$ steps, allowing the base model to form a coherent layout before object-specific prompts refine regions toward their rare concepts.

\noindent\textbf{Layer-Wise Alternating Guidance (LAG).} To further reduce early mixing between rare objects, we assign reference images to the diffusion model in a cyclic manner. Let $\{I_1, \dots, I_n\}$ denote the reference images for $n$ rare objects. At each denoising step $t$, the hidden state of layer $l$ is updated as:
   \begin{equation}
       \mathbf{h}_t^{l+1} = F_A^l\Big(\mathbf{h}_t^l, f_\mathcal{E}(c)\Big) 
                          + F_A^l\Big(\mathbf{h}_t^l, f_\phi(I_{(l \bmod n)+1})\Big),
                          \label{Eq:LAALAG}
   \end{equation}
   where $f_\phi(I_i)$ encodes the reference image of rare concepts $c_i$. 
   
\begin{figure}[t]
    \centering
    \includegraphics[width=0.7\linewidth]{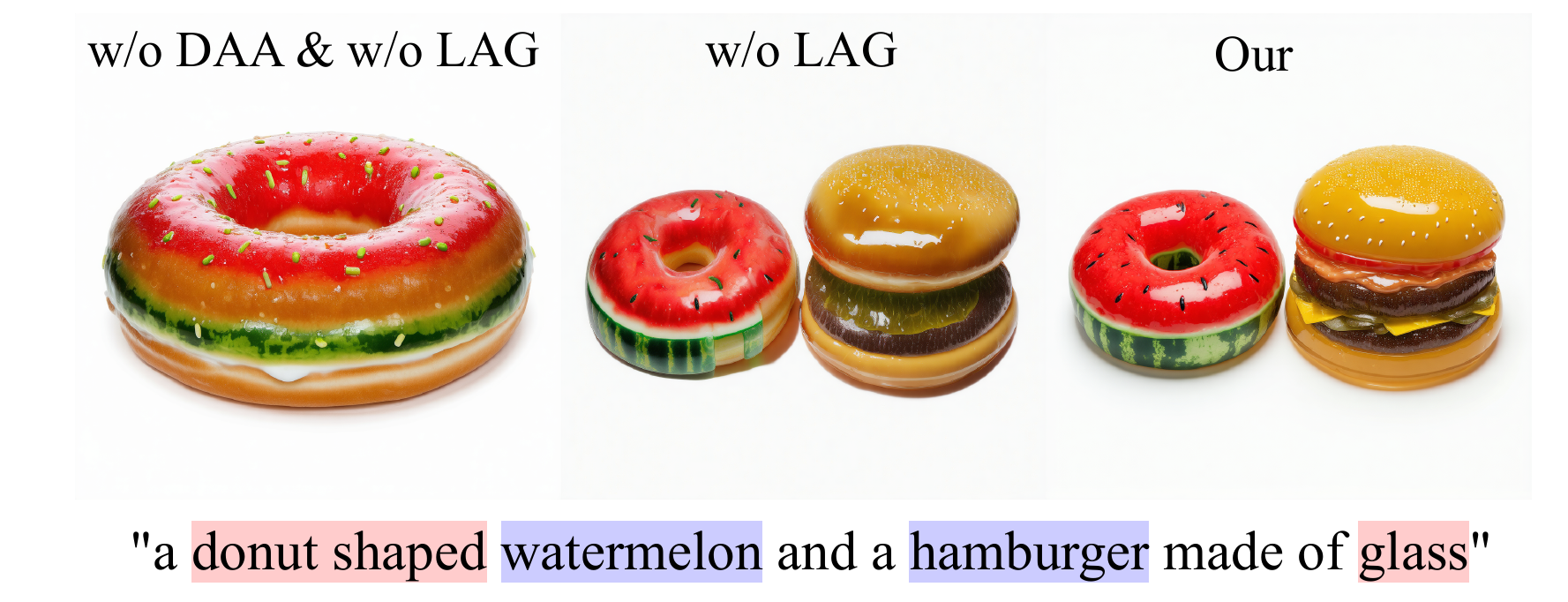}
    \caption{\textbf{Visualization of the effect of DAA and LAG.}}
    \label{fig:late_switch}
    \vspace{-8pt}
\end{figure}

   By cycling objects across layers, each rare concept is allocated dedicated capacity within distinct functional regions of the network, ensuring that their feature representations contribute independently to the denoising process. This layer-wise allocation effectively preserves distinct object identities while enabling the model to capture their complex spatial and semantic inter-relationships. As visualized in Fig.~\ref{fig:late_switch}, the synergy between DAA and LAG (formulated in~\cref{Eq:LAALAG}) provides a robust architectural defense against premature concept fusion and semantic "bleeding". Together, these mechanisms orchestrate a faithful multi-object rare concept synthesis, maintaining rigorous compositional integrity and visual coherence across diverse and challenging prompts.

\begin{table*}[t]
\centering
\resizebox{\textwidth}{!}{
\begin{tabular}{l|c|c|ccccc|ccc|c}
\hline
\multirow{2}{*}{Methods} &\multirow{2}{*}{BackBone} &\multirow{2}{*}{Venue} & \multicolumn{5}{c|}{Single}& \multicolumn{3}{c|}{Multi} & \multirow{2}{*}{Overall} \\
\cline{4-11}
 &&& Property & Shape & Texture & Action & Complex & Concat & Relation & Complex  \\ 
 \rowcolor{gray!10}
\hline
-  &SD~1.5~\cite{sd1.5}&CVPR'22& 55.0\textsuperscript{†} & 38.8\textsuperscript{†} & 33.8\textsuperscript{†} & 23.1\textsuperscript{†} & 36.9\textsuperscript{†} & 23.1\textsuperscript{†} & 24.4\textsuperscript{†} & 36.3\textsuperscript{†} & 33.9\textsuperscript{†}\\
\rowcolor{gray!10}
-   &SDXL~\cite{sdxl}&ICLR'24& 60.0\textsuperscript{†} & 56.9\textsuperscript{†} & 71.3\textsuperscript{†} & 47.5\textsuperscript{†} & 58.1\textsuperscript{†} & 39.4\textsuperscript{†} & 35.0\textsuperscript{†} & 47.5\textsuperscript{†} & 52.0\textsuperscript{†}\\
\rowcolor{gray!10}
-  &SD~3.0~\cite{sd3}&ICML'24& 49.4\textsuperscript{†} & 76.3\textsuperscript{†} & 53.1\textsuperscript{†} & 71.9\textsuperscript{†} & 65.0\textsuperscript{†} & 55.0\textsuperscript{†} & 51.2\textsuperscript{†} & 70.0\textsuperscript{†} & 61.5\textsuperscript{†}\\
\rowcolor{gray!10}
-   &FLUX~\cite{flux2024}&-& 69.4 & 78.1 & 52.5 & 65.6 & 66.9 & 63.1 & 61.3 & 81.3 & 67.3\\
\rowcolor{gray!10}
-  &SD~3.5~\cite{sd3}&ICML'24& 77.5 & 80.6 & 75.0 & 83.1 & 80.0 & 71.9 & 58.8 & 80.6 & 75.9\\
\rowcolor{orange!10}
\hline
SynGen~\cite{rassin2023linguistic} &SD~1.4~\cite{sd1.5}&NeurIPS'23& 61.3 & 54.4 & 50.6 & 45.6 & 51.2 & 32.5 & 39.4 & 40.0 & 46.9\\
\rowcolor{orange!10}
InitNO~\cite{initno}  &SD~1.4~\cite{sd1.5}&CVPR'24& 48.8 & 46.9 & 36.9 & 33.8 & 55.6 & 38.8 & 28.7 & 29.4 & 39.9\\
\rowcolor{orange!10}
InitNO~\cite{initno}  &SD~3.5~\cite{sd3}&CVPR'24& 56.9 & 62.5 & 51.2 & 67.5 & 76.2  & 56.3 & 58.1 & 75.6 & 63.0\\

\rowcolor{green!10}

%\hline
LMD~\cite{lian2023llm}    &SD~1.5~\cite{sd1.5}&TMLR'24& 23.8\textsuperscript{†} & 35.6\textsuperscript{†} & 27.5\textsuperscript{†} & 23.8\textsuperscript{†} & 35.6\textsuperscript{†} & 33.1\textsuperscript{†} & 33.4\textsuperscript{†} & 33.1\textsuperscript{†} & 30.7\textsuperscript{†}\\
\rowcolor{green!10}
RPG~\cite{yang2024mastering}    &SDXL~\cite{sdxl}&ICML'24& 33.8\textsuperscript{†} & 54.4\textsuperscript{†} & 66.3\textsuperscript{†} & 31.9\textsuperscript{†} & 37.5\textsuperscript{†} & 21.9\textsuperscript{†} & 15.6\textsuperscript{†} & 29.4\textsuperscript{†} & 36.4\textsuperscript{†}\\
\rowcolor{green!10}
ELLA~\cite{hu2024ella}   &SD~1.5~\cite{sd1.5}&arXiv'24& 33.1 & 51.2 & 65.6 & 46.9 & 60.6 & 41.3 & 48.8 & 62.5 & 51.3\\

\rowcolor{blue!10}
R2F~\cite{r2f}  &SD~3.0~\cite{sd3}&ICLR'25& 89.4\textsuperscript{†} & 79.4\textsuperscript{†} & 81.9\textsuperscript{†} & 80.0\textsuperscript{†} & 72.5\textsuperscript{†} & 70.0\textsuperscript{†} & 58.8\textsuperscript{†} & 73.8\textsuperscript{†} & 75.7\textsuperscript{†}\\
\rowcolor{blue!10}

R2F~\cite{r2f}  &SD~3.5~\cite{sd3}&ICLR'25& \underline{90.0} & \underline{82.5} & \textbf{89.4} & \underline{86.9} & \underline{80.6} & \underline{78.1} & \textbf{64.4} & \underline{81.3} &\underline{81.7}\\

\rowcolor{yellow!10}
\hline
\textbf{Our}  &SD~1.5~\cite{sd1.5}&-
& 63.1 & 55.0 & 47.5 & 42.5 & 37.5
& 33.1 & 30.6 & 43.1 & 44.0\\

\rowcolor{yellow!10}
\textbf{Our}  &SD~3.5~\cite{sd3}&-& \textbf{97.5} & \textbf{89.4} & \textbf{89.4} & \textbf{87.5} & \textbf{85.6} & \textbf{80.0} & \underline{63.1} & \textbf{85.0}&\textbf{84.7}\\
\hline
\end{tabular}
}
\caption{\textbf{Quantitative results on RareBench across various methods.} 
\textbf{Bold} values indicate the best, \underline{underlined} values the second best, and \textsuperscript{†} denotes numbers from R2F~\cite{r2f}. This comparison including different T2I metohds: \textcolor{gray!90}{T2I backbones}, 
\textcolor{orange!80!black}{attention-based T2I enhancement}, 
\textcolor{green!50!black}{LLM-grounded compositional generation}, and \textcolor{blue!80!black}{prompt-switching rare-concept generation}.}

\vspace{-20pt}
\label{tab:gpt4_single_object}
\end{table*}

\section{Experiments}
\subsection{Setup}

\noindent\textbf{Datasets.} We evaluate \abbrev on \textbf{RareBench}~\cite{r2f}, a benchmark designed for rare concept. RareBench provides 40 prompts for each rare concept type and covers five single-object concept categories. We also evaluate \abbrev on \textbf{T2I-CompBench}~\cite{huang2023t2i}, which offers a more diverse set of general compositional prompts. In particular, each attribute class (\textit{color}, \textit{shape}, \textit{texture}) contains 300 \textit{multi-object} prompts. For more detailed analysis, we extract the first object from each prompt to construct a corresponding \textit{single-object} subset, which allows us to evaluate under both single-object and multi-object settings.

\noindent\textbf{Implementation Details. } 
We implement \abbrev using SD~3.5~\cite{sd3} as the backbone diffusion model and employ publicly available IP-Adapters~\cite{ye2023ip-adapter} for compositional guidance.  All experiments are conducted with 24 denoising steps during inference, and all other hyperparameters follow the default settings of SD~3.5.  
In our method, we set $T_\text{early}=15$ and $T_\text{late}=20$ for general cases. Also we set $T_\text{early}=18$ for prompts involving relational descriptions. For CLIP-based similarity evaluation, we adopt the CLIP ViT-B/32 model.

%\paragraph{Baselines.} We compare \abbrev with a set of strong diffusion-based baselines: (1) vanilla diffusion models (SD1.5~\cite{sd1.5}, SD3.0~\cite{sd3}, SDXL~\cite{sdxl}) without adaptive guidance, (2) Rare-to-Frequent (R2F)~\cite{r2f}, (3) region-controlled methods such as SynGen~\cite{rassin2023linguistic}, (4) LLM-grounded diffusion models built on SDXL~\cite{sdxl}, including LMD~\cite{lian2023llm}, RPG~\cite{yang2024mastering}, and ELLA~\cite{hu2024ella}, (5) compositional text-to-image models PixArt-$\alpha$~\cite{pixart} and FLUX-schnell~\cite{flux}, which leverage attention-based guidance to model object relations, and (6) InitNO~\cite{initno}, which improves compositional control through initialization-based normalization.

\noindent\textbf{Evaluation Metrics. }
Following R2F~\cite{r2f}, we evaluate the generated images using GPT-4o~\cite{openai_gpt4o}-based scoring, which measures the compositional consistency between images and prompts.  
We also conduct a user study to complement the automatic evaluation.  
For the T2I-CompBench~\cite{t2i_compbench}, we follow the official setting, which adopts BLIP~\cite{blip} instead of GPT-4o as the evaluator.

\subsection{Quantitative Results}
\noindent\textbf{Performance on RareBench.} \cref{tab:gpt4_single_object}~shows that stronger backbones such as SD~3.5~\cite{sd3} and FLUX~\cite{flux2024} establish higher baselines for rare concept synthesis, confirming the impact of foundational model capacity. While LLM-grounded methods like ELLA~\cite{hu2024ella} or layout-based planners like RPG~\cite{yang2024mastering} provide structural cues, they often introduce semantic drift or over-constrain the generation, leading to lower stability compared to their original backbones~\cite{lian2023llm}. In contrast, \abbrev achieves an overall improvement of approximately 3 points over the current SOTA, R2F~\cite{r2f}. Notably, our method excels in \emph{property} and \emph{shape} categories by maintaining a feedback-driven equilibrium. Unlike R2F, which often sacrifices object identity for attribute expression, RADIANCE preserves structural integrity via its reactive ``restoring force''. Comparable \emph{texture} results suggest backbone saturation in material binding, limiting potential for further gain.

\noindent\textbf{Multi-Object and Spatial Reasoning.} In multi-object scenarios, \abbrev maintains dominance in \emph{Concat} and \emph{Complex} settings, validating the efficacy of our DAA and LAG modules in mitigating premature concept fusion. However, a slight performance ceiling in \emph{Relation} indicates that while our design effectively rebalances semantic weights, high-order spatial arrangements (e.g., \emph{left of}, \emph{behind}) remain largely dependent on the backbone’s intrinsic positional priors established in early denoising stages.

\noindent\textbf{Architecture Decoupling and Generalization.} We further validate the portability of \abbrev by applying it to heterogeneous backbones (SD~1.5 and SD~3.5) without re-tuning. While attention-based methods like InitNO~\cite{initno} are sensitive to model-specific internal maps which can distort generations during cross-architecture transfer, our framework’s reliance on image-level feedback enables more stable transfer. This robustness is further evidenced on T2I-CompBench~\cite{huang2023t2i}; unlike R2F, which may neglect base concepts in frequent-prompt regimes, \abbrev consistently improves compositional accuracy by balancing both base and rare semantic elements.

\begin{table}[t]
\small
\centering
\setlength{\tabcolsep}{3pt}
\begin{tabular}{l|ccc|ccc|c}
\hline
\multirow{2}{*}{Methods} & \multicolumn{3}{c|}{Single} & \multicolumn{3}{c|}{Multi} & \multirow{2}{*}{Overall}\\
\cline{2-7}
 & Color & Shape & Texture & Color & Shape & Texture &  \\
 \rowcolor{gray!10}
\hline
SD~1.5       & 81.8 & 75.2 & 74.0 & 34.6 & 32.8 & 38.4 & 56.1\\
 \rowcolor{gray!10}
FLUX       & 82.4 & \textbf{82.6} & \textbf{80.3} & 75.1 & 58.7 & 70.4 & \underline{74.9}\\
\rowcolor{gray!10}
SD~3.5      & 87.5 & 77.6 & 73.0 & \underline{76.2} & \underline{60.0} & \underline{71.1} & 74.2\\
\hline
\rowcolor{orange!10}
SynGen     & 86.0 & 78.5 & \underline{79.5} & 67.1 & 42.1 & 61.1 & 69.1\\
\rowcolor{green!10}
ELLA       & 86.0 & 73.5 & 76.8 & 71.9 & 47.2 & 62.6 & 69.7\\
\rowcolor{blue!10}
R2F        & \textbf{89.1} & 76.9 & 73.2 & 75.1 & 51.6 & 68.3 & 72.4\\
\hline
\rowcolor{yellow!10}
\textbf{Our}        & \underline{88.6} & \underline{79.6} & 77.0 & \textbf{77.2} & \textbf{61.8} & \textbf{71.3} & \textbf{75.9}\\
\hline
\end{tabular}
\caption{\textbf{Quantitative results on T2I-CompBench across various methods.} 
The best values are highlighted in \textbf{bold} and the second best values are \underline{underlined}.}
\vspace{-20pt}

\label{tab:t2i}
\end{table}

\begin{figure*}[th]
    \centering
    \includegraphics[width=0.9\textwidth]{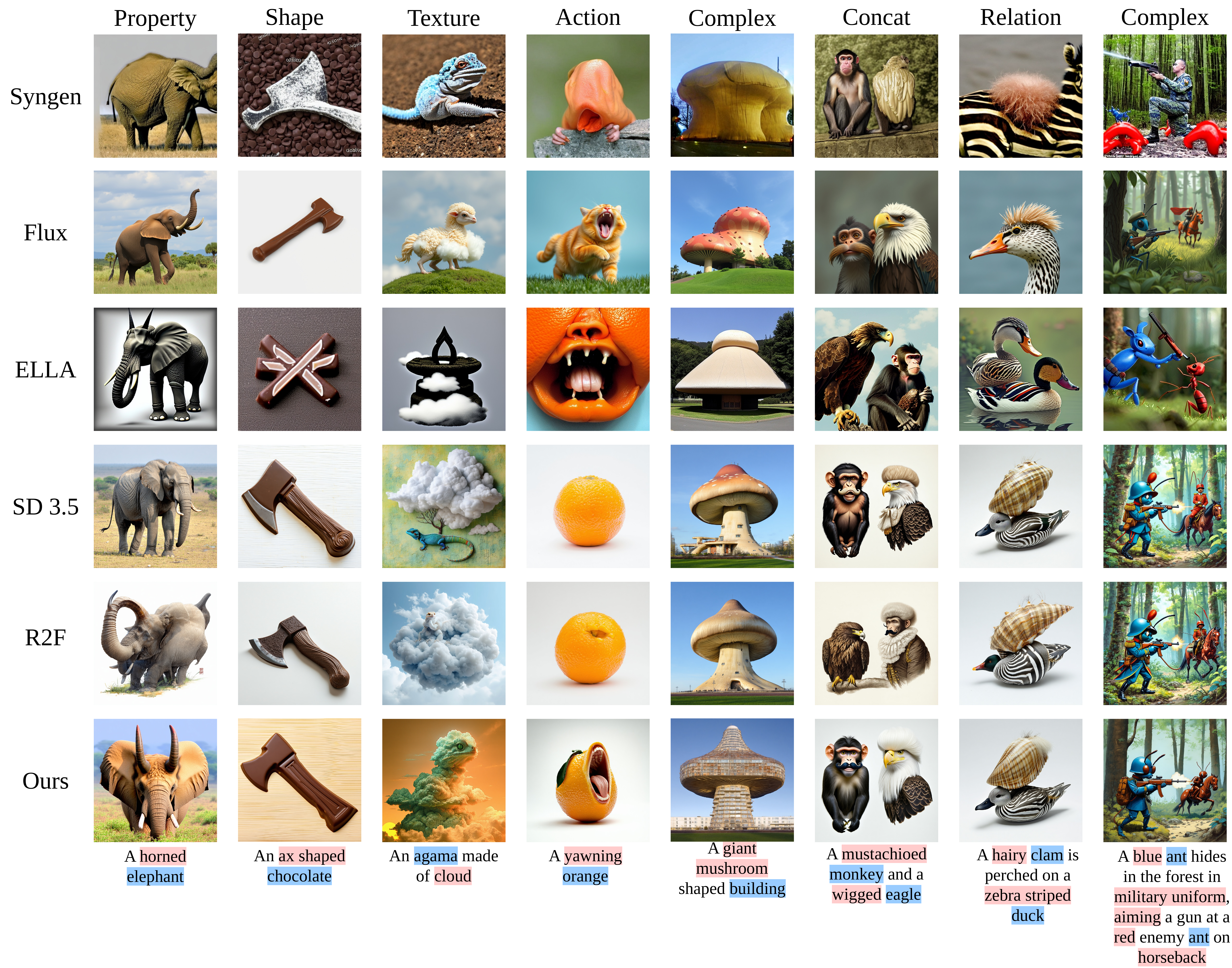}
    \vspace{-7pt}
    \caption{\textbf{Qualitative comparison on RareBench.} \abbrev~produces more accurate and coherent results compared to baselines.}
    \label{fig:qualitative}
    \vspace{-10pt}
\end{figure*}

\subsection{Qualitative Comparison}
We qualitatively compare \abbrev with representative diffusion methods and backbones (Fig.~\ref{fig:qualitative}). Original backbones often fail to generate rare attributes at all (\eg, ``a horned elephant'' in SD~3.5 and Flux). Other methods may strengthen rare attributes but let them dominate the base object (\eg, ``a giant mushroom-shaped building'' in R2F and SynGen), or fail to fuse attributes correctly with the underlying geometry (\eg, ``an ax-shaped chocolate'' in R2F and SD~3.5, where metallic regions appear on the ax head). Methods relying on older backbones perform even worse on multi-attribute or multi-object prompts. In contrast, \abbrev preserves both object identity and rare attributes, producing coherent results even under challenging multi-object compositions.

\subsection{User Preference and Inference Latency} 
We conducted a user study with 52 participants and collected 1,248 valid responses. Each response corresponds to a pair of images generated by our method and a baseline. We evaluate the results using: \textbf{Pick} (\%), the fraction of total responses that prefer our method, and \textbf{Win} (\%), the fraction of prompt category in which our method receives $>50\%$ of the votes (\cref{tab:user_and_time}). We group the prompts into two categories: (1) \textit{\textcolor{purple!70}{Basic}} (Property, Shape, Texture) and (2) \textit{\textcolor{brown!80}{Advanced}} (Action, Complex, Multi-Concat, Multi-Relation, Multi-Complex).  
Our method shows strong dominance in the \textit{\textcolor{purple!70}{Basic}} category, with small gaps between Pick and Win, indicating that users are highly sensitive to object–attribute alignment.  
For \textit{\textcolor{brown!80}{Advanced}} prompts, although our method still leads, the larger gap between Pick and Win suggests that user attention is more dispersed across multiple elements, resulting in less uniform preferences.

We also analyze the overhead of \abbrev, which adds per-step VAE decoding and CLIP monitoring. In practice, the wall-clock impact is modest (Table~\ref{tab:user_and_time}) because SD~3.5 latency is dominated by Transformer denoising, and we further overlap feedback at step $t$ with denoising at step $t{+}1$ via pipelining. 
More importantly, deployment should consider \emph{time-to-acceptable-image}: misaligned samples are typically discarded and trigger resampling. By improving compositional alignment, \abbrev\ increases the per-sample success rate and reduces the expected resampling budget, improving effective throughput (usable images per unit time) while keeping inference speed competitive.

% preamble (若你原本已有就不用重複)
% \usepackage{booktabs}
% \usepackage{multirow}
% \usepackage[table]{xcolor}

\begin{table}[t]
\centering

\begin{minipage}[t]{0.66\linewidth}
\centering
\resizebox{\linewidth}{!}{%
\begin{tabular}{l|cc|cc|cc}
\hline
\multirow{2}{*}{Comparison} & \multicolumn{2}{c|}{\cellcolor{purple!10}Basic} & \multicolumn{2}{c|}{\cellcolor{brown!15}Advanced} & \multicolumn{2}{c}{\cellcolor{green!10}Overall}\\
\cline{2-7}
 & \cellcolor{purple!5}Pick & \cellcolor{purple!15}Win
 & \cellcolor{brown!10}Pick & \cellcolor{brown!20}Win
 & \cellcolor{green!5}Pick & \cellcolor{green!15}Win\\
\hline
Our vs. SD~3.5 & \cellcolor{purple!5}95.5\% & \cellcolor{purple!15}100\% & \cellcolor{brown!10}61.5\% & \cellcolor{brown!20}60\% & \cellcolor{green!5}74.3\% & \cellcolor{green!15}75.0\%\\
Our vs. R2F    & \cellcolor{purple!5}84.6\% & \cellcolor{purple!15}100\% & \cellcolor{brown!10}60.0\% & \cellcolor{brown!20}80\% & \cellcolor{green!5}69.2\% & \cellcolor{green!15}87.5\%\\
Our vs. FLUX   & \cellcolor{purple!5}79.5\% & \cellcolor{purple!15}66.7\% & \cellcolor{brown!10}71.5\% & \cellcolor{brown!20}100\% & \cellcolor{green!5}74.5\% & \cellcolor{green!15}87.5\%\\
\hline
\end{tabular}}
\end{minipage}\hfill
\begin{minipage}[t]{0.32\linewidth}
\centering
\small
\begin{tabular}{lc}
\toprule
Method & Time (sec)\\
\midrule
SD3.5 & 7.36\\
R2F   & 7.38\\
Ours (w/o parallel) & 16.48\\
\textbf{Ours (parallel)$^{*}$} & \textbf{9.34}\\
\bottomrule
\end{tabular}
\end{minipage}
\caption{\textbf{User preference and inference latency.}
Left: user study results. Right: average wall-clock time per image. $^{*}$ denotes our default pipelined execution.}
\label{tab:user_and_time}
\vspace{-10pt}
\end{table}

\subsection{Ablation Study}
To better understand the contributions of each component in \abbrev, we conduct ablation experiments for single-object and multi-object generation. For single-object, we evaluate the impact of the Bidirectional Scale Controller (BSC) and the Feedback Guidance Scheduler (FGS) on RareBench single-object prompts. As shown in \cref{tab:ablation_single}, we consider three variants with fixed scales and a \textbf{Negative Prompt baseline}. While Negative Prompts provide a static guidance for attribute exclusion, they fail to dynamically adapt to the evolving denoising process, resulting in lower scores compared to our feedback-driven approach. Furthermore, we investigate the sensitivity of the feedback starting timestep $T_{\text{early}}$. As $T_{\text{early}}$ aligns with the critical ``sweet spot'' where the diffusion process transitions from low-frequency structure to high-frequency details, we observe that performance peaks at $T_{\text{early}}{=}15$ (marked as $^*$). The results show stability across neighboring steps, validating that our design captures the intrinsic generative priors of diffusion models rather than relying on narrow hyperparameter tuning. Overall, these results confirm that dynamic, feedback-driven scaling is essential for faithful single-object generation.

For multi-object prompts, we examine the effect of our two strategies for preventing premature fusion: \textit{Delayed Adapter Activation (DAA)} and \textit{Layer-wise Alternating Guidance (LAG)}. As shown in \cref{tab:ablation_multi}, disabling either component causes clear performance drops. In \textit{Concat} and \textit{Relation} settings with fewer objects, DAA is more critical for avoiding early merging of concepts. As the number of objects grows and scenes become more complex, LAG brings larger gains by strengthening the representation of each object. Together, these results show that both mechanisms are necessary to maintain object identities and accurate attribute–object bindings in multi-object synthesis.

% Single-object ablation
\begin{table}[t]
\small
\centering
\setlength{\tabcolsep}{3pt} % 調整欄間距
\begin{tabular}{l|ccccc|c}
\hline
\multirow{2}{*}{Methods}   & \multicolumn{5}{c|}{Single} & \multirow{2}{*}{Overall} \\
\cline{2-6}
 & Property & Shape & Texture & Action & Complex & \\
\cline{1-6}
\hline
Fixed $s_o$     & 46.3 & 65.6 & 71.9 & 35.0 & 63.1 & 56.4  \\ 
Fixed $s_a$      & 66.3 & \underline{84.4} & 57.5 & 55.0 & 71.9 & 67.0 \\ 
Fixed both       & 83.8 & 51.2 & 82.5 & 78.8 & 68.8 & 73.0 \\  
w/o BSC         & \underline{93.8} & 83.8 & \underline{88.8} & \underline{86.3} & \underline{81.3} & \underline{86.8}\\ 
Negative prompt& 87.5 &84.4 &87.5 &83.1 &74.4& 83.4\\
$T_{\text{early}}{=}13$ &94.4&88.1&88.1&\textbf{88.1}&86.9&89.1\\
$T_{\text{early}}{=}15${*} & \textbf{97.5} & \textbf{89.4} & \textbf{89.4} & 87.5 & 85.6 & \textbf{89.9}\\  
$T_{\text{early}}{=}17$ & 95.6&87.5&86.3&83.8&\textbf{87.5}& 88.1\\ 
\hline
\end{tabular}
\caption{\textbf{Ablation results on RareBench.} We evaluate the Bidirectional Scale Controller (BSC) against fixed-scale variants and a negative prompt baseline. The impact of the feedback start time $T_{\text{early}}$ is also shown, with $15^*$ indicating our default setting used in the final model.}
\label{tab:ablation_single}
\vspace{-20pt}
\end{table}

% Multi-object ablation
\begin{table}[t]
\centering
\resizebox{0.4\textwidth}{!}{ % 改小一點寬度
\begin{tabular}{l|ccc|c}
\hline
\multirow{2}{*}{Methods} & \multicolumn{3}{c|}{Multi} & \multirow{2}{*}{Overall} \\
\cline{2-4} 
 & Concat & Relation & Complex & \\
\cline{1-5}
\hline
w/o DAA  & 65.6 & 40.6 & \underline{81.9} &  62.7  \\ 
w/o LAG  & \underline{76.2} & \underline{58.1} & 78.8 &  \underline{71.0}  \\  
Our   & \textbf{80.0} & \textbf{63.1} & \textbf{85.0} & \textbf{76.0} \\  
\hline
\end{tabular}
}
\caption{\textbf{Ablation study for multi-object prompts.} Each variant disables one of the two strategies to prevent premature fusion of multiple reference images.}
\label{tab:ablation_multi}
\vspace{-24pt}
\end{table}

\section{Conclusion}
We propose \abbrev, a training-free, feedback-driven framework designed to resolve the inherent compositional imbalances in rare concept generation. By monitoring similarity and applying bidirectional scaling, we can dynamically balances object and attribute guidance during sampling, which leads to more semantically coherent images across diverse prompts. Extensive experiments on RareBench and T2I-CompBench show that \abbrev consistently improves compositional alignment and rare attribute fidelity over strong baselines, while preserving the visual quality of the underlying diffusion models.

\noindent\textbf{Limitations and Future Work.}
While \abbrev significantly improves the semantic fidelity of rare attribute binding , its current design prioritizes conceptual alignment over global spatial arrangement. Consequently, in scenes requiring precise spatial reasoning among numerous objects, the final layout remains primarily constrained by the backbone's intrinsic geometric priors. We view this as a strategic trade-off: \abbrev provides a robust semantic foundation that is a prerequisite for complex composition.  Future work will investigate the integration of our closed-loop semantic controller with spatially-aware guidance and cross-object interaction modeling to further enhance compositional controllability in densely populated or spatially complex environments.

\clearpage

\section*{Acknowledgements}
This work was partially supported by the National Science and Technology Council, Taiwan (Grants: NSTC-112-2221-E-A49-094-MY3 and NSTC-112-2221-E-A49-059-MY3, NSTC-114-2640-E-A49-011).

% ---- Bibliography ----
%
% BibTeX users should specify bibliography style 'splncs04'.
% References will then be sorted and formatted in the correct style.
%
\bibliographystyle{splncs04}
\bibliography{main}

\appendix

\end{document}